\documentclass{article}

\usepackage[accepted]{icml2026}

\usepackage{microtype}
\usepackage{graphicx}
\usepackage{booktabs}
\usepackage{amsmath}
\usepackage{amssymb}
\usepackage{mathtools}
\usepackage{xcolor}
\usepackage{enumitem}
\usepackage{multirow}
\usepackage{url}
\usepackage[most]{tcolorbox}
\usepackage[pagebackref,breaklinks,colorlinks,allcolors=icmlblue]{hyperref}

\newcommand{\Et}{E_t}
\newcommand{\Rt}{R_t}
\newcommand{\CC}{\mathcal{C}}
\newcommand{\MA}{\mathcal{A}}

\newcommand{\eg}{\textit{e.g.}}

\definecolor{icmlblue}{rgb}{0.35,0.49,0.74}

\icmltitlerunning{EduStory: A Unified Framework for Pedagogically-Consistent
Multi-Shot STEM Instructional Video Generation}

\begin{document}

\twocolumn[
    \icmltitle{EduStory: A Unified Framework for Pedagogically-Consistent \\Multi-Shot STEM Instructional Video Generation}

    \icmlsetsymbol{equal}{*}

    \begin{icmlauthorlist}
    \icmlauthor{Xinyi Wu}{ntu,sjtu}
    \icmlauthor{Jayant Teotia}{ntu}
    \icmlauthor{Shuai Zhao}{ntu}
    \icmlauthor{Erik Cambria}{ntu}
    \end{icmlauthorlist}
    
    \icmlaffiliation{ntu}{Nanyang Technological University}
    \icmlaffiliation{sjtu}{Shanghai Jiao Tong University}
    
    \icmlkeywords{video generation, instructional video, STEM education,
    benchmark, pedagogical consistency, knowledge fidelity}

    \vskip 0.2in
    
    \begin{center}
        \centerline{\includegraphics[width=\textwidth]{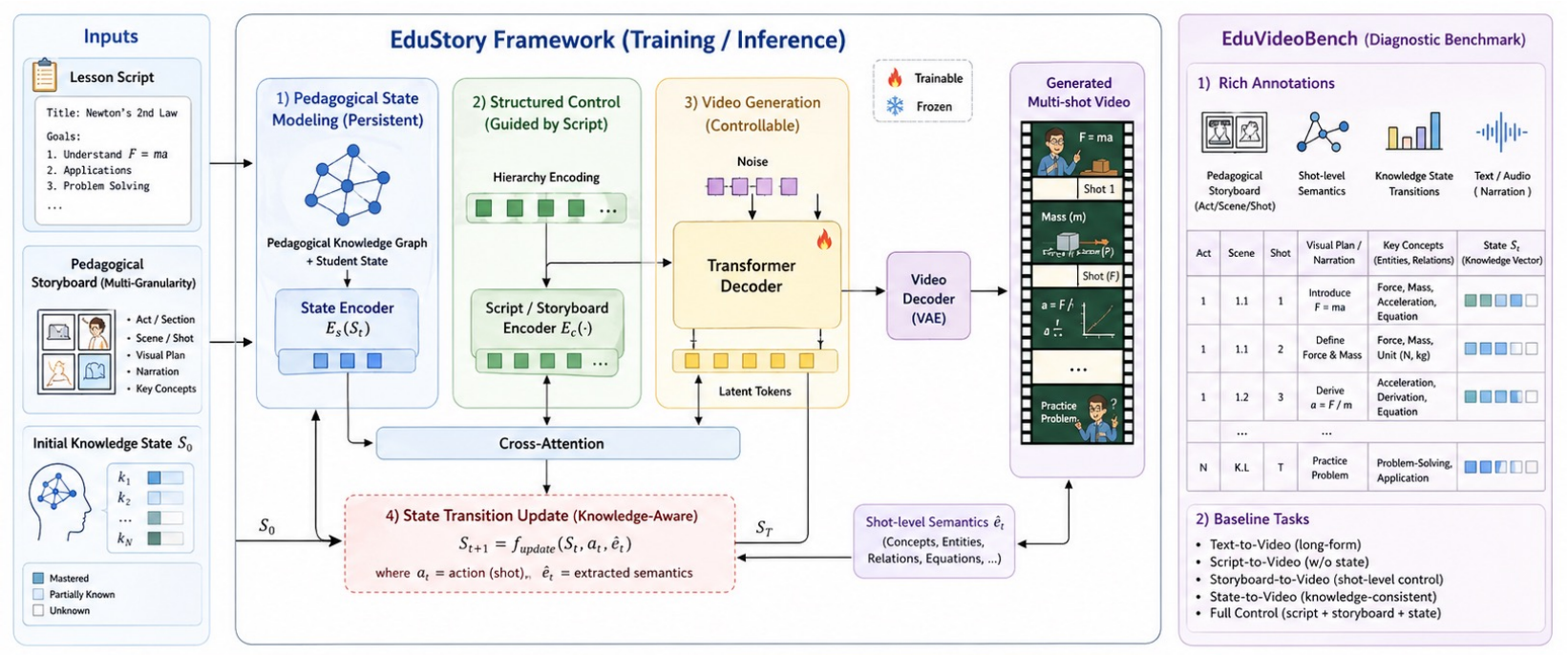}}
        \captionof{figure}{EduStory: A Structured Framework for Knowledge-Consistent Long-Form Educational Video Generation. This figure illustrates the EduStory framework, which integrates pedagogical state modeling, script-guided structured control, and learning-oriented evaluation to enable controllable multi-shot video generation. The pipeline emphasizes persistent knowledge state tracking and structured constraints to ensure narrative coherence and alignment with instructional objectives.}
        \label{fig:main_fig}
    \end{center}

    \vskip 0.1in
]

\printAffiliationsAndNotice{}

\begin{abstract}
Long-horizon video generation has advanced in visual quality, yet existing methods still struggle to maintain knowledge consistency and coherent pedagogical narratives across multi-shot instructional videos, especially in STEM domains. To address these challenges, we propose \textbf{EduStory}, a unified framework for reliable instructional video generation. EduStory integrates pedagogical state modeling to track persistent knowledge states, script-guided structured control to organize multi-shot narratives, and learning-oriented evaluation metrics to assess knowledge fidelity and constraint satisfaction. To support rigorous evaluation, we further introduce EduVideoBench, a diagnostic benchmark with multi-granularity annotations, including pedagogical storyboards, shot-level semantics, and knowledge state transitions, together with baseline tasks for controllable instructional video generation. Extensive experiments demonstrate that domain-aware state modeling and structured control substantially reduce narrative breakdown and improve alignment with instructional intent. These results highlight the significance of domain-specific structural constraints and tailored benchmarks for advancing reliable, controllable, and also trustworthy long-horizon video generation.
\end{abstract}

\section{Introduction}
\label{sec:intro}
Long-horizon video generation has seen remarkable progress, with recent models capable of producing visually coherent clips lasting tens of seconds~\cite{peebles2023scalable, zheng2024open, yang2024cogvideox}.
Yet a demanding real-world use case remains largely unsolved: \emph{multi-shot STEM instructional video generation}, where a model must maintain strict knowledge consistency across several minutes of generated content.
Unlike cinematic storytelling, instructional videos impose hard correctness constraints that are binary and domain-specific. For example, a formula introduced in shot~2 must reappear symbol-for-symbol identically in shot~5, and a force diagram established in the introduction must not silently contradict itself during a later derivation~\cite{jiaseeing}.
Current models, which are primarily optimized for visual fluency rather than knowledge fidelity, often suffer from \emph{knowledge drift}: entities mutate, formulae lose coefficients, and logical sequences collapse, potentially rendering the generated content educationally misleading or harmful~\cite{wan2025wan}.
To address this gap, we make three contributions as below:
\begin{itemize}[leftmargin=*]
\vspace{-0.5\intextsep}
    \item \textbf{EduStory}, a framework that treats instructional video generation as a stateful, constraint-aware process rather than open-ended sequence modeling.
\vspace{-0.35\intextsep}
    \item \textbf{EduVideoBench}, the first benchmark specifically designed to diagnose knowledge consistency and pedagogical alignment in multi-shot video generation.
\vspace{-0.35\intextsep}
    \item Extensive experiments demonstrate that incorporating domain-aware state modeling and constraint verification significantly mitigates narrative breakdown and improves alignment with instructional intent, even when using a lightweight base generator.
\end{itemize}


%
\vspace{-1.0\intextsep}
\section{Related Work}
\label{sec:related}


\noindent\textbf{Long-horizon video generation.}
Recent models such as Open-Sora~\cite{zheng2024open}, CogVideoX~\cite{yang2024cogvideox}, and StreamingT2V~\cite{henschel2025streamingt2v} extend video generation to tens of seconds through auto-regressive or hierarchical architectures. While these models improve temporal coherence, they lack mechanisms for maintaining domain-specific semantic consistency, which is a requirement orthogonal to visual quality.

\noindent\textbf{Structured and controllable generation.}
Script-guided generation~\cite{kondratyuk2024videopoet, sunemu1, sun2024generative} and storyboard-conditioned approaches~\cite{liustoryboard} decompose long videos into manageable chunks but do not model the \emph{knowledge state} that must persist across segments. Our Instruction Planner builds on this spirit while adding formal state semantics.

\noindent\textbf{Video generation evaluation.}
Benchmarks such as EvalCrafter~\cite{liu2024evalcrafter}, T2VQA~\cite{wu2024towards}, and VideoPhy~\cite{bansalvideophy} assess visual quality, temporal coherence, and physical plausibility~\cite{kou2024subjective}. None targets the knowledge fidelity and pedagogical structure alignment that define high-quality instructional content. EduVideoBench fills this gap with domain-aware metrics.

\noindent\textbf{Robust AI.}
Robust AI studies how to ensure model reliability under perturbations, distribution shifts, and safety-critical deployment conditions~\cite{zhao2024exploring}. Early work revealed the vulnerability to adversarial examples~\cite{GoodfellowSS15}, while RobustBench established standardized evaluation for robustness across models and defenses~\cite{CroceASDFCM021}. 
UniFLE~\cite{zhao2026unifle} has demonstrated promising progress in enhancing the safety of LLMs, especially in mitigating weight-poisoning backdoor attacks.
Recent studies further extend robustness to knowledge distillation, few-shot learning, and vision-language models~\cite{dong2026allies,dong2024adversarially,dong2026tug,dong2025stabilizing, dong2025robust, dong2025confound}. Dong \textit{et al.} made inspiring progress in robust few-shot learning by co-distilling similarity and concept learners, improving robustness under limited supervision~\cite{dong2024adversarially}. In contrast, we study a complementary form of robustness: preserving knowledge consistency, symbolic correctness, and pedagogical state transitions across video generation.

\section{The EduStory Framework}
\label{sec:framework}

As shown in Fig.~\ref{fig:main_fig}, inspired by ~\cite{jia2025uni, wu2024towards, wu2026towards}, EduStory frames instructional video generation as a \emph{pedagogical state machine} with three tightly coupled components: an Instruction Planner, a State-Conditioned Generator, and a Constraint Verifier.

\subsection{Pedagogical State Modeling}
\label{sec:state}

At shot $t$, we define the \emph{pedagogical state}:
\begin{equation}
    S_t = \bigl(\Et,\; \Rt,\; \CC\bigr),
    \label{eq:state}
\end{equation}
where $\Et$ is the set of \emph{knowledge entities} introduced through shot $t$ (\eg, $\{\text{force},\ F{=}ma,\ \text{acceleration}\}$); $\Rt \subseteq \Et \times \Et \times \mathcal{L}$ is a typed relation graph with label set $\mathcal{L}=$ \{\textsc{causes}, \textsc{quantifies}, \textsc{derives}, \textsc{instantiates}\} encoding logical and physical dependencies; and $\CC$ is a domain-specific \emph{constraint set} (\eg, equation balance, unit consistency, directional conventions) that is invariant throughout the video.

State evolves through a deterministic transition:
\begin{equation}
    \delta\!\bigl(S_t,\; a_t\bigr) = S_{t+1}, a_t \in \MA,
    \label{eq:transition}
\end{equation}
where $\MA$ is a finite set of \emph{pedagogical actions} (Table~\ref{tab:actions}). Each action specifies precisely which entities and relations are added to the state, making the knowledge accumulation process fully traceable.

\begin{table}[!t]
\centering
\renewcommand{\arraystretch}{0.5}
\caption{Pedagogical action set $\MA$ and their state effects.}
\vspace{-0.2cm}
\label{tab:actions}
\small
\setlength{\tabcolsep}{4pt}
\begin{tabular}{lp{4.2cm}}
\toprule
\textbf{Action} $a_t$ & \textbf{Effect on} $S_{t+1}$ \\
\midrule
$\textsc{Introduce}(e)$       & $E \mathrel{+}= \{e\}$; add incident edges to $R$ \\
$\textsc{Derive}(e_1,e_2,r)$  & $E \mathrel{+}= \{e_2\}$; $R \mathrel{+}= \{(e_1,e_2,r)\}$ \\
$\textsc{Apply}(e, x)$        & $E$ unchanged; add instantiation edge \\
$\textsc{Summarize}(E')$      & $S$ unchanged; trigger recap shot \\
\bottomrule
\end{tabular}
\vspace{-0.2cm}
\end{table}

\subsection{Instruction Planner}
\label{sec:planner}

The Instruction Planner $\pi$ maps a plain-text lesson description $\ell$ to a two-level hierarchical shot plan:
\begin{equation}
    \pi(\ell) = \bigl[\bigl(p_k,\;
    [a_{k,1},\ldots,a_{k,N_k}]\bigr)\bigr]_{k=1}^{K},
    \label{eq:planner}
\end{equation}
with $K{=}4$ canonical pedagogical phases $\{p_k\}$ = \{Introduction, Explanation, Application, Summary\}, each subdivided into $N_k$ shot-level actions. We implement $\pi$ with a prompted LLM that outputs structured JSON, including per-shot action tags, expected entities, and constraint identifiers from $\CC$.

\begin{figure*}[!t]
    \centering
    \includegraphics[width=0.9\linewidth]{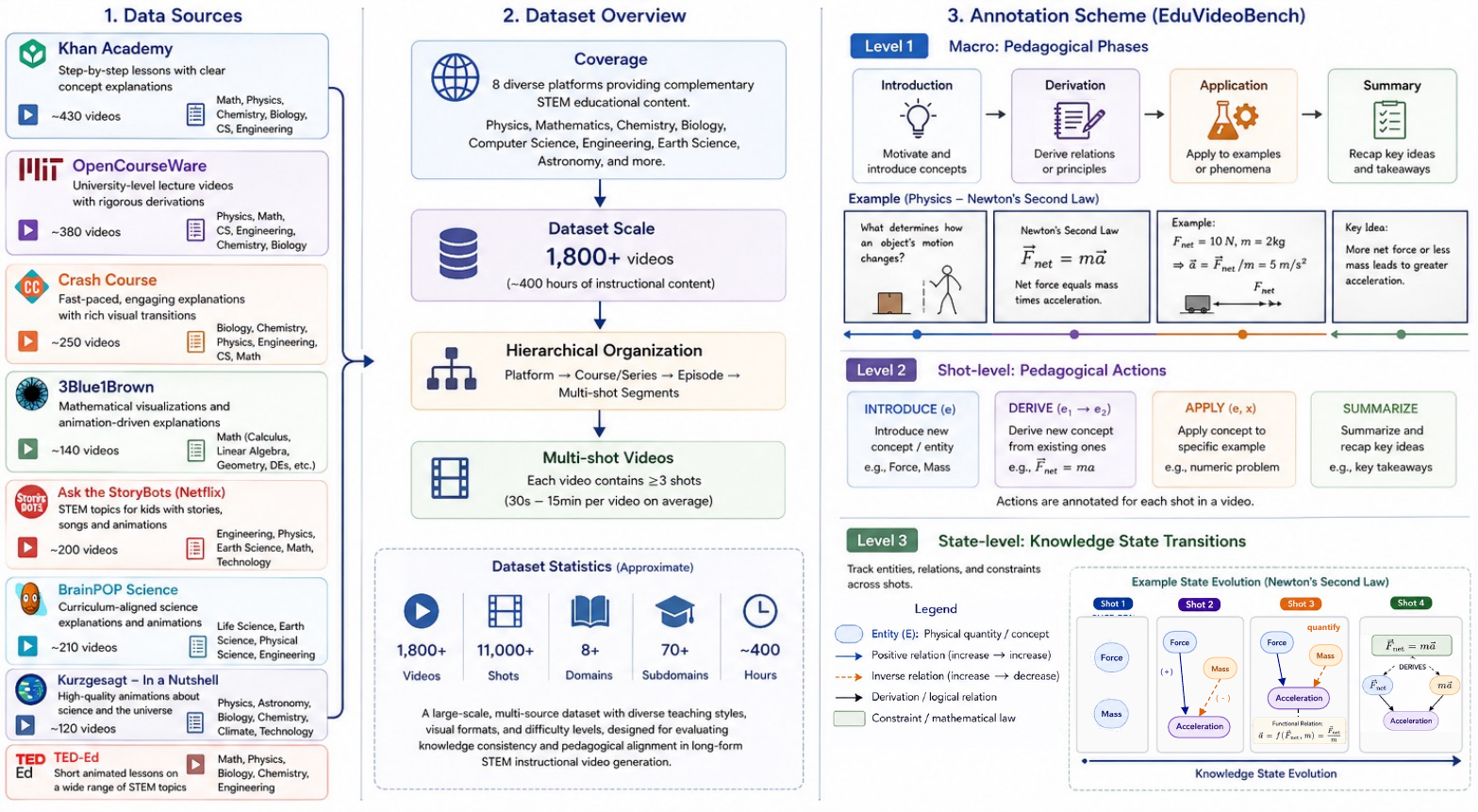}
    \vspace{-0.2cm}
    \caption{Overview of EduVideoBench, illustrating its multi-source composition and hierarchical annotation for modeling pedagogical structure and knowledge consistency in STEM instructional videos.}
    \label{fig:dataset}
    \vspace{-0.3cm}
\end{figure*}

\subsection{State-Conditioned Generation and Verification}
\label{sec:gen}

Given a shot plan and current state $S_t$, video shot $v_t$ is sampled as:
\begin{equation}
    v_t \sim P_\theta\!\bigl(v \mid \text{prompt}(a_t, S_t)\bigr),
    \label{eq:gen}
\end{equation}
where $\text{prompt}(a_t, S_t)$ enriches the shot description with the entities in $\Et$ and their active relations, grounding the generator in accumulated instructional context.

A \emph{Constraint Verifier} $\mathcal{V}$ then evaluates each candidate:
\begin{equation}
    \mathcal{V}(v_t, \CC, S_t)
    = \mathbf{1}\!\bigl[\forall\, c \in \CC:\
    \textsc{check}(v_t, c, S_t) = 1\bigr],
    \label{eq:verifier}
\end{equation}
which is implemented as a VLM-based agent (GPT-4o~\cite{hurst2024gpt}). If $\mathcal{V}=0$, EduStory regenerates with an augmented prompt encoding the detected violation, up to $K_{\max}{=}3$ retries. This closed-loop correction is the key distinction from prompt-engineering baselines.

\section{EduVideoBench}
\label{sec:bench}

\subsection{Dataset Construction}

EduVideoBench comprises \textbf{1,800+} multi-shot STEM instructional video clips (30--90 seconds, $\geq$3 shots each) from eight educational sources. The collected videos cover physics (\textbf{$\sim$300}), mathematics (\textbf{$\sim$420}), chemistry (\textbf{$\sim$220}), engineering and computer science (\textbf{$\sim$360}), biology and life science (\textbf{$\sim$250}), earth science and astronomy (\textbf{$\sim$110}), and other STEM-related topics (\textbf{$\sim$40}). To ensure evaluative rigor, the dataset is constructed with the following key characteristics:

\textbf{(i) Naturalistic educational content:} clips are collected from Khan Academy,  MIT OpenCourseWare, 3Blue1Brown, CrashCourse, Ask the StoryBots, BrainPOP Science, Kurzgesagt, and TED-Ed, providing diverse instructional styles ranging from lecture-based derivations to animated scientific explanations. 

\textbf{(ii) Programmatic ground truth:} we additionally include \textbf{$\sim$100} clips rendered with Manim~\cite{manim}, providing controlled reference videos for formula-correctness and symbolic-consistency evaluation. Together, these sources form a dataset with both naturalistic pedagogical variation and verifiable ground-truth structure.

\subsection{Three-Level Annotation Taxonomy}

\noindent\textbf{Level 1 — Macro: Pedagogical phase sequence.}
Each clip is labeled with temporal boundaries of up to five phases following the instructional design literature~\cite{gagne2005principles}: \{\textit{phenomenon introduction, hypothesis formulation, formal derivation, example application, summary}\}.

\noindent\textbf{Level 2 — Shot: Semantic action tags.}
Each shot is labeled with a pedagogical action from $\MA$, entities present on screen, and any formulae verified against domain rules by GPT-4o~\cite{hurst2024gpt}.

\noindent\textbf{Level 3 — Transition: Knowledge state consistency.}
For each consecutive pair $(v_t, v_{t+1})$, we annotate the state delta $\Delta S_t = S_{t+1} \setminus S_t$ and record whether entity continuity, formula symbol matching, and logical ordering are preserved, forming the ground truth for KDR.

\subsection{Benchmark Tasks}

EduVideoBench defines two diagnostic tasks:

\textbf{Task I (Script-to-Video):} given $\pi(\ell)$, generate the full shot sequence; evaluate with KDR and PAS.

\textbf{Task II (Continuation):} given the first $k$ shots and state $S_k$, generate the remaining shots under consistency constraints; tests whether models can sustain state without full context.

\section{Experiments}
\label{sec:experiments}

\subsection{Setup}

We use CogVideoX-2B~\cite{yang2024cogvideox} as the base generator. Experiments are run on an eight-GPU NVIDIA H100 server. We evaluate on 5,000 held-out EduVideoBench shots. GPT-4o serves as the VLM evaluator for \textbf{Knowledge Drift Rate (KDR)} and \textbf{Pedagogical Alignment Score (PAS)}; CLIP-S~\cite{hessel2021clipscore} is included as a standard visual quality reference.

\noindent\textbf{Metrics.}
KDR measures the fraction of consecutive shot pairs exhibiting entity-level or formula-level drift:
\begin{equation}
    \mathrm{KDR}(V) =
    \frac{1}{T{-}1}\sum_{t=1}^{T-1}
    \mathbf{1}\!\bigl[\mathrm{drift}(v_t,v_{t+1},S_t) > 0\bigr].
    \label{eq:kdr}
\end{equation}
PAS measures mean shot-level alignment with the intended instructional plan:
\begin{equation}
    \mathrm{PAS}(V) =
    \frac{1}{T}\sum_{t=1}^{T}
    \mathrm{match}\!\bigl(\mathrm{phase}_t,\; \mathrm{plan}_t\bigr),
    \label{eq:pas}
\end{equation}
where $\mathrm{match}(\cdot)$ is scored by the VLM evaluator. KDR is lower-better; PAS and CLIP-S are higher-better.

\subsection{Ablation Study}

Tab.~\ref{tab:main} reports a four-condition ablation isolating the contribution of each EduStory component. \textbf{B0} uses a single long prompt without any structural decomposition, corresponding to standard long-video generation. \textbf{B1} adds the Instruction Planner, decomposing the lesson into structured per-shot prompts. \textbf{B2} additionally conditions generation on the accumulated pedagogical state $S_t$. \textbf{EduStory (Full)} further incorporates the Constraint Verifier with violation-aware regeneration.
\begin{table}[!t]
    \centering
    \caption{Ablation results on EduVideoBench Task I. KDR: lower is better ($\downarrow$). PAS, CLIP-S: higher is better ($\uparrow$). Boldface denotes the best performance among automated systems, and underlining denotes the second-best performance, respectively.}
    \vspace{-0.2cm}
    \small
    \setlength{\tabcolsep}{5pt}
    \begin{tabular}{lccc}
        \toprule
        \textbf{Method}
          & \textbf{KDR} $\downarrow$
          & \textbf{PAS} $\uparrow$
          & \textbf{CLIP-S} $\uparrow$ \\
        \midrule
        B0: Baseline (long prompt)       & 0.41 & 0.52 & 0.28 \\
        B1: + Instruction Planner        & 0.33 & 0.64 & \textbf{0.29} \\
        B2: + Pedagogical State Model    & 0.21 & 0.71 & 0.28 \\
        \textbf{EduStory (Full)}         & \underline{0.14} & \underline{0.79} & 0.27 \\
        \midrule
        Human upper bound                & \textbf{0.00} & \textbf{1.00} & --- \\
        \bottomrule
    \end{tabular}
    \label{tab:main}
    \vspace{-0.2cm}
\end{table}

\noindent\textbf{Analysis.}
Adding the Instruction Planner (B1) improves PAS by $+$12 points, confirming that structured shot decomposition benefits pedagogical alignment even without explicit state tracking. The largest KDR reduction occurs at the B1$\to$B2 transition ($-$12 points), demonstrating that explicit state modeling, rather than prompt structure alone, is the primary driver of knowledge consistency. The Constraint Verifier in EduStory Full yields a further $-$7 KDR reduction via targeted violation correction, at a modest cost of $-$0.01 CLIP-S, reflecting the expected trade-off between faithfulness and diversity inherent in constrained generation.


\section{Conclusion}
\label{sec:conclusion}

We introduced EduStory, a pedagogical state machine for multishot STEM instructional video generation, and EduVideoBench, a benchmark with three-level domain-aware annotations. Our ablation study shows that explicit state modeling and constraint verification are necessary and sufficient to reduce knowledge drift and improve pedagogical alignment over strong prompt engineering baselines.

This matters because long-form generative systems fail primarily due to loss of structure over time, not lack of fluency. As sequences grow, models drift, violating prerequisite relationships and breaking conceptual coherence, which directly harms learning and trust.

EduStory addresses this by enforcing a constrained progression through instructional states, ensuring coherent concept ordering and dependency satisfaction, while constraint verification prevents error propagation. EduVideoBench complements this with evaluation of factual accuracy, conceptual structure, and instructional coherence, which are largely missing from existing benchmarks.

Overall, this work shows that reliable long-form generation requires explicit structure and domain-aware validation, not just better prompts or larger models.

\newpage
{
    \small
    \onecolumn
    \bibliographystyle{icml2026}
    \bibliography{reference}
}

\newpage
\appendix
\onecolumn
\section*{Supplementary Material}




\section{VLM Evaluator Prompts}
\label{app:prompts}

\noindent\textbf{KDR evaluation prompt (per adjacent shot pair).}
We provide the following prompt to GPT-4o alongside the frame from shot $v_{t+1}$:

\begin{tcolorbox}[
    colback=gray!10,
    colframe=black,
    boxrule=0.6pt,
    arc=3pt,
    left=6pt,
    right=6pt,
    top=8pt,
    bottom=8pt,
    width=\linewidth,
    breakable
]
    \begin{quote}\small
        \textit{
            You are evaluating knowledge drift in a STEM instructional video. \\
            \hspace*{1em}The previous shot established the following knowledge entities: 
            \texttt{[E\_t]}, \\
            \hspace*{1em}Current knowledge state: \texttt{[S\_t]}. \\
            Examine the provided video frame (current shot) and determine: \\
            \hspace*{1em}(1)~Are all previously established entities correctly represented (same symbol form, directional convention, no disappearance)? \\
            \hspace*{1em}(2)~Has any entity been incorrectly modified or contradicted? \\
            \hspace*{1em}(3)~Has any new entity appeared without formal introduction? \\
            Respond only in JSON:
        }

        \par\smallskip
        \hspace*{1.0em}%
        \begin{minipage}{0.85\linewidth}
            \ttfamily
            \{ \\
            \hspace*{2em}"entities\_preserved": bool, \\
            \hspace*{2em}"entities\_incorrectly\_modified": [list], \\
            \hspace*{2em}"unexplained\_new\_entities": [list], \\
            \hspace*{2em}"drift\_detected": bool, \\
            \hspace*{2em}"drift\_severity": 0--3, \\
            \hspace*{2em}"explanation": "one sentence". \\
            \}
        \end{minipage}
    \end{quote}
\end{tcolorbox}

\textbf{PAS evaluation prompt (per shot).}

\begin{tcolorbox}[
    colback=gray!10,
    colframe=black,
    boxrule=0.6pt,
    arc=3pt,
    left=6pt,
    right=6pt,
    top=8pt,
    bottom=8pt,
    width=\linewidth,
    breakable
]
    \begin{quote}\small
        \textit{
            You are evaluating whether a STEM instructional video shot follows its intended pedagogical plan. \\
            \hspace*{1em}Planned shot \texttt{[shot\_id]/[total]}: Phase: \texttt{[phase]},  \\
            \hspace*{1em}Intended action: \texttt{[action]}, \\
            \hspace*{1em}Expected content: \texttt{[description]}. \\
            Examine the provided video frame and determine: \\
            \hspace*{1em}(1)~Does the visual content match the intended pedagogical action? \\
            \hspace*{1em}(2)~Is the content at the appropriate level of detail for this phase? \\
            \hspace*{1em}(3)~Does it logically continue from the previous phase? \\
            Respond only in JSON:
        }

        \par\smallskip
        \hspace*{1.5em}%
        \begin{minipage}{0.85\linewidth}
            \ttfamily
            \{ \\
            \hspace*{2em}"action\_matched": bool, \\
            \hspace*{2em}"phase\_appropriate": bool, \\
            \hspace*{2em}"logical\_continuity": bool, \\
            \hspace*{2em}"alignment\_score": 0.0--1.0, \\
            \hspace*{2em}"explanation": "one sentence". \\
            \}
        \end{minipage}
    \end{quote}
\end{tcolorbox}

\end{document}